\documentclass[letterpaper, 10pt, conference]{ieeeconf}  

\IEEEoverridecommandlockouts                              

\usepackage{color}
\usepackage{amsmath, amssymb, amscd, amsfonts} 
\usepackage{graphics} 
\usepackage{graphicx}
\usepackage{hyperref}
\usepackage{tabularx}
\usepackage{multirow}
\usepackage{colortbl}
\usepackage{hhline}
\usepackage{booktabs}
\usepackage{lipsum}
\usepackage{cite}
\usepackage{float}
\usepackage{subfig}
\usepackage{caption}
\usepackage{algorithm}
\usepackage{algorithmic}
\usepackage{bbding}

\UseRawInputEncoding

\newcommand{\method}{\textit{DriveSceneGen}}

\title{\LARGE \bf
    DriveSceneGen: Generating Diverse and Realistic Driving Scenarios from Scratch
}

\author{
    Shuo Sun$^{\dag}$, Zekai Gu$^{\dag}$, Tianchen Sun$^{\dag}$, Jiawei Sun, Chengran Yuan, \\ Yuhang Han, Dongen Li, and Marcelo H. Ang Jr.
    \thanks{
        $^{\dag}$Equal Contributions.
    }
    \thanks{
        All Authors are with the Advanced Robotics Centre, National University of Singapore, Singapore 119077 (e-mail: \{shuo.sun, zekai.gu, tianchen.sun, sunjiawei, chengran.yuan, yuhang\_han, li.dongen\}@u.nus.edu; mpeangh@nus.edu.sg).
    }%
    \thanks{
        Project page: \href{https://ss47816.github.io/DriveSceneGen/}{https://ss47816.github.io/DriveSceneGen/}.
    }%
}

\begin{document}

\maketitle
\thispagestyle{empty}
\pagestyle{empty}


\begin{abstract}

Realistic and diverse traffic scenarios in large quantities are crucial for the development and validation of autonomous driving systems. However, owing to numerous difficulties in the data collection process and the reliance on intensive annotations, real-world datasets lack sufficient quantity and diversity to support the increasing demand for data. This work introduces \textit{DriveSceneGen}, a data-driven driving scenario generation method that learns from the real-world driving dataset and generates entire dynamic driving scenarios from scratch. \method~is able to generate novel driving scenarios that align with real-world data distributions with high fidelity and diversity. Experimental results on 5k generated scenarios highlight the generation quality, diversity, and scalability compared to real-world datasets. To the best of our knowledge, \method~is the first method that generates novel driving scenarios involving both static map elements and dynamic traffic participants from scratch. 
\end{abstract}



\section{INTRODUCTION}


In recent years, data-driven approaches have garnered increasing popularity in the autonomous driving research community. 
Many of the latest achievements in this field are powered by deep learning models trained on large amounts of data collected from real-world driving scenarios. 
For this class of methods, having large amounts of different scenarios is paramount for their performance and generalization capability. 
Particularly for the decision-making and planning modules in autonomous systems, it is important to ensure safety and sufficient capability to handle challenging scenarios by training and validating the system on a wide variety of driving scenarios. 
For this reason, the demand for more driving scenarios has been rapidly growing over the years. 
As part of this endeavor, numerous research institutions and autonomous vehicle companies have released a range of public driving datasets\cite{Waymo2021, nuPlan2021, Argoverse2019, Argoverse2021}. 
However, despite the fact that these are real-world datasets collected from various locations covering a large number of scenarios, several notable limitations still persist. 

\subsubsection{Limited quantity and variety}
In terms of data quantity, most existing datasets only contain about 300 to 500 hours, 290 to 1750 km of driving data. In comparison, an average American driver drives 21,688 km annually \cite{USDT2022}. The data quantity is still not satisfactory compared to human statistics. In terms of data diversity, most datasets were only collected in a few US cities, limited by the available testing areas. Within each dataset, the number of unique scenes, i.e., geo-locations with distinct map typologies, is also limited, further constraining the generalizability of the dataset. 

\begin{figure}[t]
    \centering
    \includegraphics[width=0.95\linewidth]{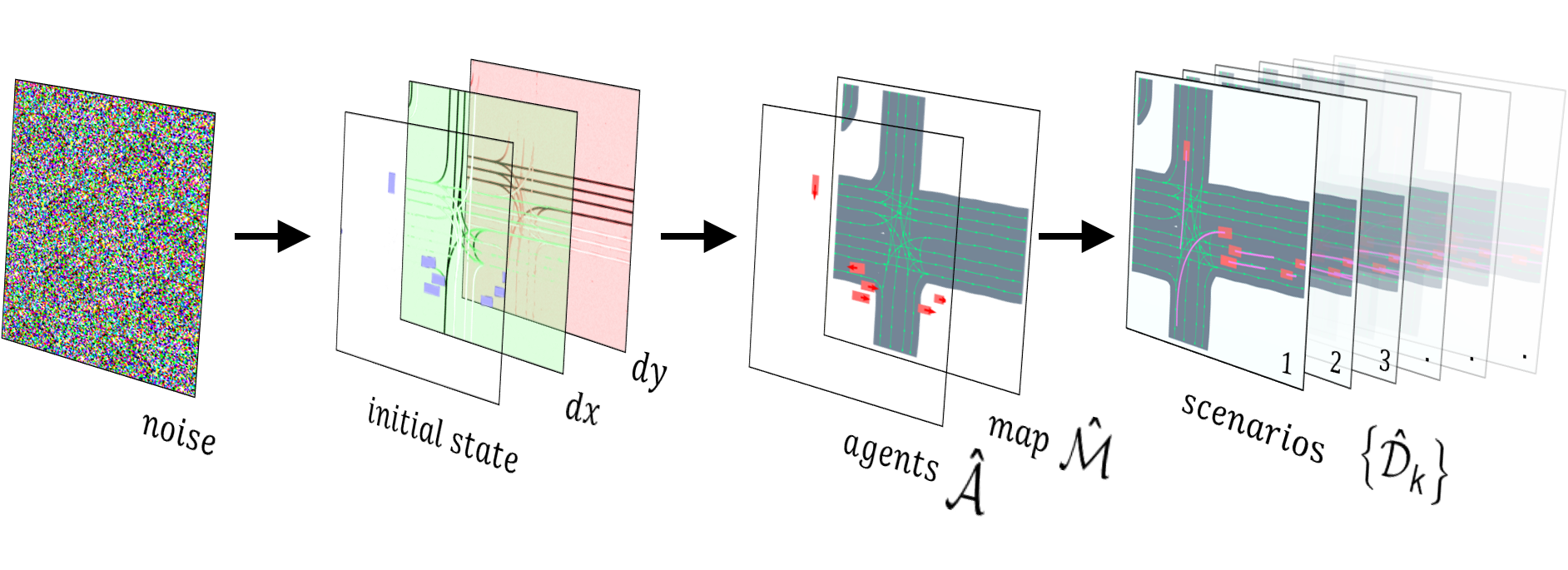}
    \caption{
        An illustration of a set of generated scenarios and its components at the various stages of \method~pipeline.
    }
    \label{figure-abstract}
\end{figure}

\subsubsection{Non-deterministic nature of agents' driving behaviors} 
It is worth noting that the joint distribution of all traffic participants' behaviors is inherently non-deterministic. Starting from the same initial scene, an unlimited number of possible futures could unfold. At every time step, each traffic participant receives new partial observations and makes individual judgments upon every smallest nuance occurring in the scene, resulting in drastically different driving scenarios in the end. However, real-world datasets could only capture the one that was developed at the time of collection, which can cause a significant loss of data diversity. 

\subsubsection{Expensive and Time-consuming}
One of the major limitations of real-world datasets is that the collection process requires specialized equipment and licensing, often making the acquisition process extremely expensive. The post-processing and labeling of the collected scenario data also tend to be labor-intensive and time-costly, further limiting the availability of data to most researchers and institutes. 

\textbf{Motivation and Contributions.}
With the understanding of these limitations of existing real-world datasets and the rising need for more data, we wish to leverage the latest generative models to learn the real-world distribution of various driving scenarios and then generate novel scenarios from this learned distribution. The contributions of this paper can be summarized in fourfold:
\begin{itemize}
    \item We propose the first method for generating novel driving scenarios from scratch, including both static lane maps and dynamic traffic agents, following a generation-simulation two-stage pipeline. 
    \item We introduce a diffusion model in the generation stage to generate a realistic birds-eye-view (BEV) feature map of the scenario, detailing both static map elements and dynamic agents' initial states, which are then converted to its vectorized data form by a graph-based vectorization method. 
    \item We repurpose a trajectory prediction model in the simulation stage to predict the multi-modal behaviors of each agent and form multiple joint predictions as different possible future scenarios, conditioned on the generated map and agents' initial states. 
    \item Experimental evaluations on the 5$k$ scenarios generated by our proposed method demonstrate that our method is able to achieve high fidelity and diversity compared to the 70k ground truth scenarios. 
\end{itemize}



\section{RELATED WORKS}

\subsection{Diffusion models}

Denoising Diffusion Probabilistic Models (DDPM))\cite{diffusion2015, DDPM2020}, often simply referred to as diffusion models, is a new class of generative models. These models demonstrate high generation quality in image-related tasks such as image generation\cite{lantentdiffusion}, inpainting\cite{inpainting}, denoising, etc.
At the same time, diffusion models have also gained success in a wide range of other domains, including sequence generation\cite{DiffuSeq2022, Diffwave2021}, decision-making,  planning\cite{DiffsionDecisionMaking2022, PlanningwithDiffusion2022} and character animation\cite{hmd}. Experimental evidence also showed that training with synthetic data generated by diffusion models can improve task performance on tasks such as image classification\cite{DiffusionClassification2023}. Building on prior works that demonstrated the model's ability to produce high-quality samples in various domains, we are the first to apply diffusion models for the generation of autonomous driving datasets.

\subsection{Simulation Networks and Map Generation}

A typical driving scenario mainly consists of two layers of information: a static map layer detailing the lane network and a dynamic agent layer that records the traffic agents and their corresponding driving trajectories. 
Many previous works have attempted to generate diverse agent trajectories based on a given real map. 
Works including \cite{SceneGen2021, SimNet2021, TrafficGen2022} focused on sampling agent states from existing scenarios, then erasing and respawning the agents based on agent state distribution. 
Trafficgen\cite{TrafficGen2022} encoded both lane and agent data into a graph structure and summarized agent states using a Gaussian Mixture Model (GMM) and employing a Multi-Layer Perceptron (MLP) to model and simulate the distribution of agent states.
SimNet\cite{SimNet2021} constructed the dynamic state of agents as a Markov Process and further learned both the state distribution and transition function using a Neural Network.
However, these methods are only able to generate dynamic agents with a reliance on a given static map as the condition. 

One existing work that attempted static map generation is HDMapGen\cite{mi2021hdmapgen}, which autoregressively generated static lane structures as hierarchical graphs. However, the local details of the generated maps are not sufficiently realistic. The model is not capable of generating complex and diverse road networks, limiting its practical applications. 

To bridge the gap in the existing research between static map generation and dynamic agent simulation, we propose an entire driving scenario generation technique that includes static and dynamic elements, enhancing the diversity of scenarios and maintaining the consistency between static maps and dynamic agents.


\begin{figure*}[t]
    \centering
    \includegraphics[width=0.95\linewidth]{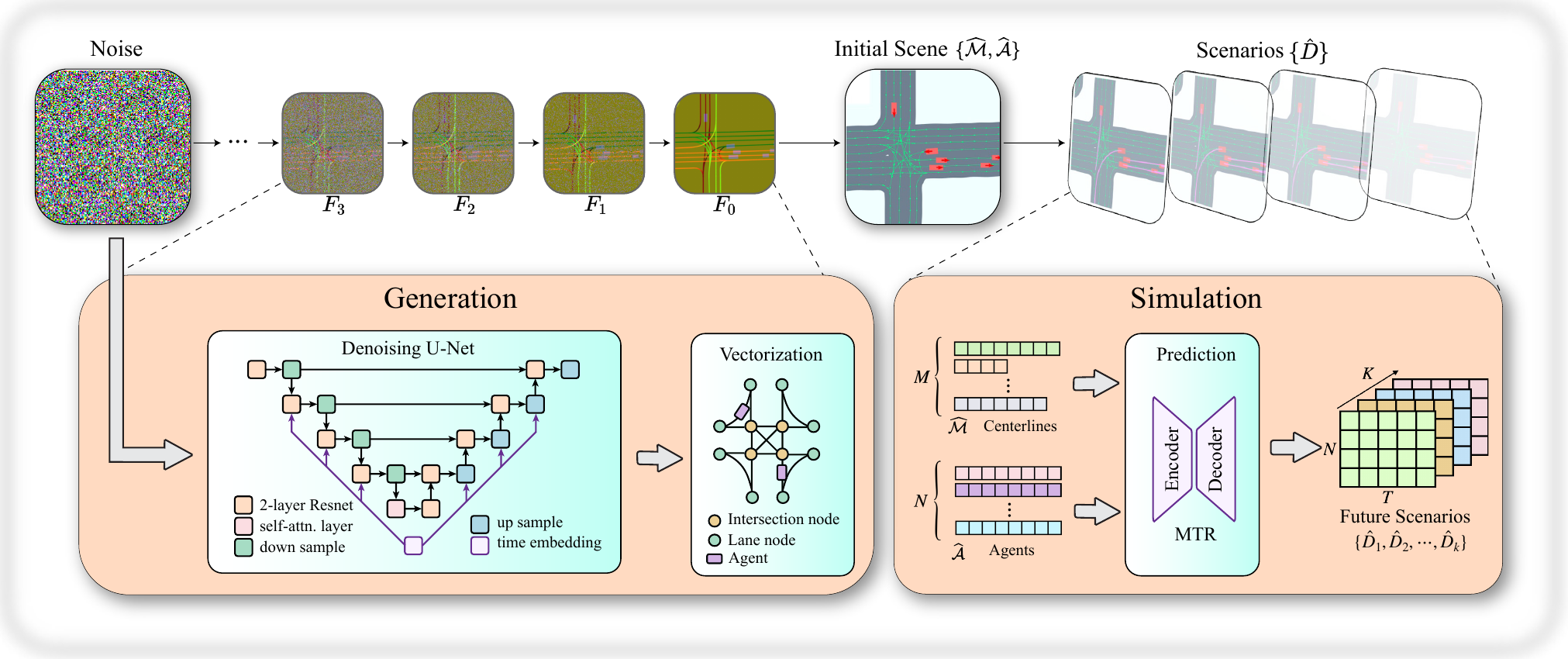}
    \caption{ An overview of the \method~pipeline. The pipeline consists of two stages: a generation stage and a simulation stage. In the generation stage, a diffusion model is employed to generate a rasterized Birds-Eye-View (BEV) representation of the initial scene of the driving scenario, which is then decoded by a graph-based vectorization method. In the simulation stage, the vectorized representation of the scenario is consumed by a simulation network as the initial scene to predict multi-modal joint distributions of the generated agents' future trajectories, with each distribution representing a distinctly possible outcome from this same initial scene. }
    \label{figure-pipeline}
\end{figure*}

\section{METHODOLOGY}

\subsection{Overview of the Proposed Method} 

The proposed method mainly consists of two stages: a generation stage and a simulation stage, as shown in Fig. \ref{figure-pipeline}. 

First, in the generation stage, we employ a diffusion model to generate a rasterized Birds-Eye-View (BEV) representation of the initial scene of the driving scenario, which is then transformed into a vector format by a graph-based vectorization algorithm. 
Second, in the simulation stage, the vectorized representation of the scenario is consumed by a simulation network as the initial scene to predict multi-modal joint distributions of the generated agents' future trajectories, with each distribution representing a distinct possible outcome from this same initial scene.

\subsection{Generation}

\subsubsection{Scenario Representation} A typical driving scenario $D$ can be divided into two major components: a static map layer $\mathcal{M}$ detailing the lane network and a dynamic agent layer $\mathcal{A}$ that records a set of traffic agents and their corresponding driving trajectories. The map $\mathcal{M}$ consists of a set of lanelets (or lanes), where each lane can be described by an imaginary polyline $\mathcal{P}$ as its centerline, which is essentially a set of waypoints $p \in \mathbb{R}^{2}$ connected according to the driving direction. Over the entire scenario time span $\mathbf{T}$, the map information is assumed to be invariant, while each agent $A_{i} \in \mathcal{A}$ is updated at every time step $t$. An agent's states over this time span can be represented by a bounding box $b \in \mathbb{R}^{2}$ containing the length and width of the vehicle, and a trajectory $\pi$, which consists of a set of vehicle states $\{s_{t} \mid 0 \leq t \leq \mathbf{T} \}$. Each state $s_{t} = [x, y, \theta, v]^{T}$ is a vector describing the vehicle x, y coordinates, heading, and velocity at time step $t$. 

For the generation task, we define a rasterization function $f \colon D(\mathcal{M}, \mathcal{A}) \rightarrow \mathbf{F}$, which encodes the above-mentioned map and agent data of a single driving scenario within a local region of $R$ meters into one rasterized BEV feature map $\mathbf{F} \in \mathbb{R}^{W \times H \times C}$, where $W$ and $H$ denote the image width and height, and $C$ denotes the channels. At each coordinate location $(x, y)$, the value $C_{x,y}^{i}$ at each channel $i$ is used to encode a different type of information. 

We propose an encoding strategy that augments the current centerlines with explicit directional information to help the model better learn the topological connection among them. For each waypoint $p = [x, y]^{T}$ on the centerline, we append an additional directional vector $\Delta p = [dx, dy]^{T}$, which points from waypoint $p_{i}$ to $p_{i+1}$, to the existing coordinate vector. Then, the centerlines are represented in the feature map with the $dx, dy$ encoded as values at two channels, respectively: 

\begin{equation}
        [C_{x, y}^{1}, C_{x, y}^{2}]^{T} = 0.5 \times \left( 1 + \frac{ \Delta p }{ \| \Delta p \|_{2} } \right) \\
\label{equation-dxdy}
\end{equation}
The directional vectors are assumed to be $[0, 0]^{T}$ for non-lane areas. For each scenario, a transform $\mathbf{T} \colon (x,y) \rightarrow (i,j)$ is computed to map all real-world coordinates into a pixel location on the feature map. 

Meanwhile, each agent's initial state is represented by a rectangle bounding box $bbox$ according to its position and heading on the third channel of the feature map $C_{x, y}^{3}$, with the values within the bounding box region representing the vehicle's velocity $v$. A background value of zero is assigned for regions in the feature map not occupied by any agent instead:

\begin{equation}
    C_{x,y}^{3} = \left\{
    \begin{array}{cl}
        0.5 \times \left( 1 + \frac{v}{v_{max}} \right) & \quad \textrm{if } (x, y) \subseteq bbox \\
        0 & \quad \textrm{otherwise}
    \end{array}\right.
\label{equation-vehicle}
\end{equation}
where $v_{max}$ represents the maximum agent velocity allowed, which is held constant for all scenarios. 

\subsubsection{Diffusion Process}

With each driving scenario $D$ being encoded into a rasterized format $\mathbf{F}$ with the strategies described above, we define a forward diffusion process $\mathbf{F}_{0}, \mathbf{F}_{1}, \cdots, \mathbf{F}_{t}$, where $T$ denotes the maximum diffusion steps. The purpose of this forward diffusion process is to destroy the information in $\mathbf{F}$ by gradually adding Gaussian noise and ultimately turning it into a complete Gaussian noise. On the contrary, we use a diffusion model to learn a reverse process $ \mathbf{F}_{t}, \mathbf{F}_{t-1}, \cdots, \mathbf{F}_{0} $, gradually removing uncertainty from the complete Gaussian noise to generate a scenario. The transition process of the forward and reverse diffusion process satisfies the properties of the Markov chain.

For the diffusion process, the posterior distribution of $\mathbf{F}_{0}, \mathbf{F}_{1}, \cdots, \mathbf{F}_{t}$ are be defined as:
\begin{equation}
    \begin{aligned}
        q\left(\mathbf{F}_{1: T} \mid \mathbf{F}_{0}\right) &:=\prod_{t=1}^{T} q\left(\mathbf{F}_{t} \mid \mathbf{F}_{t-1}\right) \\
        q\left(\mathbf{F}_{t} \mid \mathbf{F}_{t-1}\right) &:=\mathcal{N}\left(\mathbf{y}_{t} ; \sqrt{1-\beta_{t}} \mathbf{F}_{t-1}, \beta_{t} \mathbf{I}\right)
    \end{aligned}
\end{equation}
where $\beta_{1},\beta_{2}...\beta_{t}$ are fixed variance schedulers that control the scale of the injected noise. 

The result at any noise level $t$ in the diffusion process can be calculated as \cite{DDPM2020}:
\begin{equation}
    q\left(\mathbf{F}_{t} \mid \mathbf{F}_{0}\right):=\mathcal{N}\left(\mathbf{F}_{t} ; \sqrt{\overline{\alpha}_{t}} \mathbf{F}_{0},\left(1-\overline{\alpha}_{t}\right) \mathbf{I}\right)
\end{equation}
where $\alpha_{t}=1-\beta_{t}$ and $\overline{\alpha}_{t}=\prod_{t=1}^{t} \alpha_{t}$. 

When $t$ is large enough, we can approximately consider $\mathbf{F}_{T} \sim \mathcal{N}(\mathbf{0}, \mathbf{I})$. That indicates the process in which scenario $\mathbf{F}$ is gradually broken into Gaussian noise. Then, the process that generates a new scenario from Gaussian noise can be defined as a reverse diffusion process. This process can be described as：

\begin{equation}
    \begin{aligned}
        p_{\theta}\left(\mathbf{F}_{0: T} \right)&:=p\left(\mathbf{F}_{T}\right) \prod_{t=1}^{T} p_{\theta}\left(\mathbf{F}_{t-1} \mid \mathbf{F}_{t}, \right) \\
        p_{\theta}\left(\mathbf{F}_{t-1} \mid \mathbf{F}_{t}\right) &:=\mathcal{N}\left(\mathbf{F}_{t-1} ; \boldsymbol{\mu}_{\theta}\left(\mathbf{F}_{t}, t\right) ; \boldsymbol{\Sigma}_{\theta}\left(\mathbf{F}_{t}, t\right)\right)
    \end{aligned}
\end{equation}
where $p\left(\mathbf{F}_{T}\right)\sim \mathcal{N}(\mathbf{0}, \mathbf{I})$ denotes the initial noise that sample from Gaussian distribution. $\theta$ denotes the parameters of diffusion model, which acquire by training with existing driving scenarios. 

The variance term of the Gaussian transition can be set as $\boldsymbol{\Sigma}_{\theta}\left(\mathbf{F}_{T}, k\right)=\sigma_{t}^{2} \mathbf{I}=\beta_{t} \mathbf{I}$. As shown in \cite{DDPM2020}, $\boldsymbol{\mu}_{\theta}\left(\mathbf{F}_{t}, t\right)$ is obtained by calculating $F_t$ and model output, which follows the formula below：

\begin{equation}
    \boldsymbol{\mu}_{\theta}\left(\mathbf{F}_{t}, t\right)=\frac{1}{\sqrt{\alpha_{k}}}\left(\mathbf{F}_{t}-\frac{\beta_{t}}{\sqrt{1-\overline{\alpha}_{t}}} \epsilon_{\theta}\left(\mathbf{F}_{t}, t\right)\right)
\end{equation}


\subsubsection{Architecture}
For this task, a U-Net\cite{unet2015} model is employed as the backbone in the DDPM model to learn the driving scenario distribution through the reverse diffusion process and predict the noise at each diffusion step. 
As shown in Fig. \ref{figure-pipeline}, the U-Net model comprises four down and up blocks, together with a middle block. Each down block and up block consists of two ResNet layers and a Down/Up sample layer, while the middle block is made up of a self-attention layer and two ResNet layers. 

\subsubsection{Training}
The purpose of training is to maximize the log-likelihood of the generated scenario given the ground truth 
$\mathbb{E}\left[\log p_{\theta}\left(\mathbf{F}_{(0)}\right)\right]$. Since the exact log-likelihood is intractable, we follow the standard Evidence Lower Bound (ELBO) maximization method and minimize the KL Divergence.
\begin{equation}
    \begin{aligned}
        \mathcal{L} &= \mathbb{E}_{q, t}\left[D_{KL}\left(q\left(\mathbf{F}_{t-1} \mid \mathbf{F}_{t}, \mathbf{F}_{0}\right)|| p_{\theta}\left(\mathbf{F}_{t-1} \mid \mathbf{F}_{t}\right)\right)\right] \\ 
        &= \mathbb{E}_{q, t}\left[D_{KL}\left(\mathcal{N}\left(\mathbf{F}_{t-1} ; \boldsymbol{\mu}_{q}, \boldsymbol{\Sigma}_{q}(t) \| \mathcal{N}\left(\mathbf{F}_{t-1} ; \boldsymbol{\mu}_{\theta}, \boldsymbol{\Sigma}_{t}\right)\right)\right]\right. \\
        &= \mathbb{E}_{q, t}\left[\left\|\boldsymbol{\mu}_{\theta}-\boldsymbol{\mu}_{q}\right\|_{2}^{2}\right]
    \end{aligned}
\end{equation}

By utilizing the reparameterization trick, this is equivalent to obtaining the diffusion model by minimizing the expected $L2$ denoising error for training samples $x$ under any Gaussian noise level $T$.  The corresponding optimization problem becomes:

\begin{equation}
    L_{MSE}(\theta)=\mathbb{E}_{\theta, \mathbf{F}_{0}, t}\left\|\epsilon-\theta\left(\mathbf{F}_{t}, t\right)\right\|
\end{equation}
where $\epsilon \sim \mathcal{N}(0, \mathbf{I})$, $\mathbf{F}_t=\sqrt{\overline{\alpha}_{t}} \mathbf{F}_{0}+\sqrt{1-\overline{\alpha}_{t}} \theta\left(\mathbf{F}_{t}, t\right), t \in 1,2, \cdots, T$, $\theta\left(\mathbf{F}_{t}, t\right)$ denotes the model predict noise.

\subsubsection{Inference}
Once the model is trained, we are able to generate an entire new scenario by a Gaussian noise $\mathbf{F}_{T}\sim \mathcal{N}(\mathbf{0}, \mathbf{I})$. During the reverse process, DDPM \cite{DDPM2020} sampling technique repeatedly denoises $\mathbf{F}_{t}$ to $\mathbf{F}_{1}$ by using equation below for $T$ steps:

\begin{equation}
    \mathbf{F}_{t-1}=\frac{1}{\sqrt{\alpha_{t}}}\left(\mathbf{F}_{t}-\frac{\beta_{t}}{\sqrt{1-\overline{\alpha}_{t}}} \theta\left(\mathbf{F}_{t}, t\right)\right)+\sqrt{\beta_{t}} \mathbf{U}
\end{equation}
where $\mathbf{U}\sim \mathcal{N}(\mathbf{0}, \mathbf{I})$ and $\theta$ is the trained network whose inputs include the previous step’s generation $\mathbf{F}_t$ and step $t$.

\subsubsection{Vectorization}
Once the diffusion model generates a new scenario sample, a vectorization process is applied to decode the scenario information from our customized feature map representation to its vector form $f^{-1} \colon \hat{\mathbf{F}} \rightarrow \hat{D}$. It is worth highlighting that, in some applications, the generated scenarios in their current form can be directly used by methods that use raster data as inputs. However, for a more general purpose of use, we explicitly reconstruct the vectorized data $\hat{D}$ from the feature map $\hat{\mathbf{F}}$. 

\begin{algorithm}[h]
\caption{Graph-based Vectorization}
\label{algo-vector}
\begin{algorithmic}[1]
    \algsetup{linenosize=\scriptsize}
    \small
    \REQUIRE ~~\\ 
        Generated feature map $\hat{\mathbf{F}}$. \\
    \ENSURE ~~\\ 
        Vectorized Lane Graph $\vec{G}$. \\
        \STATE Construct $\hat{\mathbf{F}}_{skel} \gets \text{skeletonize}(\hat{\mathbf{F}})$
        \STATE Find $\mathcal{E}, \mathcal{V} \gets \text{extract\_edges\_vertices}(\hat{\mathbf{F}}_{skel})$
        \STATE Construct undirected graph $G\{\mathcal{V}, \mathcal{E}\}$
        \STATE Compute $\Delta \Bar{p}_{V} \gets [C_{x, y}^{1}, C_{x, y}^{2}]^{T} \times 2 - 1$, $\forall V \in \mathcal{V}$
        \STATE Find $\mathcal{V}^{t} \gets \{ V \in \mathcal{V} \mid \text{degree}(V) = 1 \} $ 
        \STATE Find $\mathcal{V}^{b} \gets \{ \text{neighbor}(V) \mid V \in \mathcal{V}^{t} \} $ 
        \STATE Label $\text{type}(V, \Delta \Bar{p}_{V}) \gets \begin{cases}
                \text{entry} \\
                \text{exit} 
            \end{cases}$, $\forall V \in \mathcal{V}^{t} \cup \mathcal{V}^{b}$
        \STATE Construct directed graph $\vec{G}\{\emptyset, \emptyset\}$
        \FORALL{$ \{V_{i}^{t}, V_{i}^{b} \} \in \{ \mathcal{V}^{t}, \mathcal{V}^{t} \} $}
            \STATE $\vec{G} \text{.add\_edge} (\vec{E}(V_{i}^{t}, V_{i}^{b}))$
            \STATE $G \text{.remove\_edge} (E(V_{i}^{t}, V_{i}^{b}))$
            \STATE $G \text{.remove\_vertex} (V_{i}^{t})$
        \ENDFOR
        \STATE Find $\mathcal{V}^{in} \gets \{ V \in G \mid \text{type}(V) = \text{entry} \} $ 
        \STATE Find $\mathcal{V}^{out} \gets \{ V \in G \mid \text{type}(V) = \text{exit} \} $ 
        \FORALL{$ V_{i}^{in} \in \mathcal{V}^{in} $}
            \FORALL{$ V_{j}^{out} \in \mathcal{V}^{out} $}
                \IF{$ \exists Path(V_{i}^{in}, V_{j}^{out}) \in G $}
                    \STATE Fit $B_{i,j} \gets \text{b\'ezier\_curve}(Path(V_{i}^{in}, V_{j}^{out}))$
                    \IF{$ \text{IoU}(B_{i,j}, Path(V_{i}^{in}, V_{j}^{out})) \geq 0.5 $ \AND \\ 
                        $ \text{max\_curvature}(B_{i,j}) \leq k_{thresh} $}
                        \STATE $\vec{G} \text{.add\_edge} (\vec{E}(V_{i}^{in}, V_{j}^{out}))$
                    \ENDIF
                \ENDIF
            \ENDFOR
        \ENDFOR
        \RETURN $\vec{G}$
\end{algorithmic}
\end{algorithm}

\begin{figure}[h]
    \centering
    \includegraphics[width=0.9\linewidth]{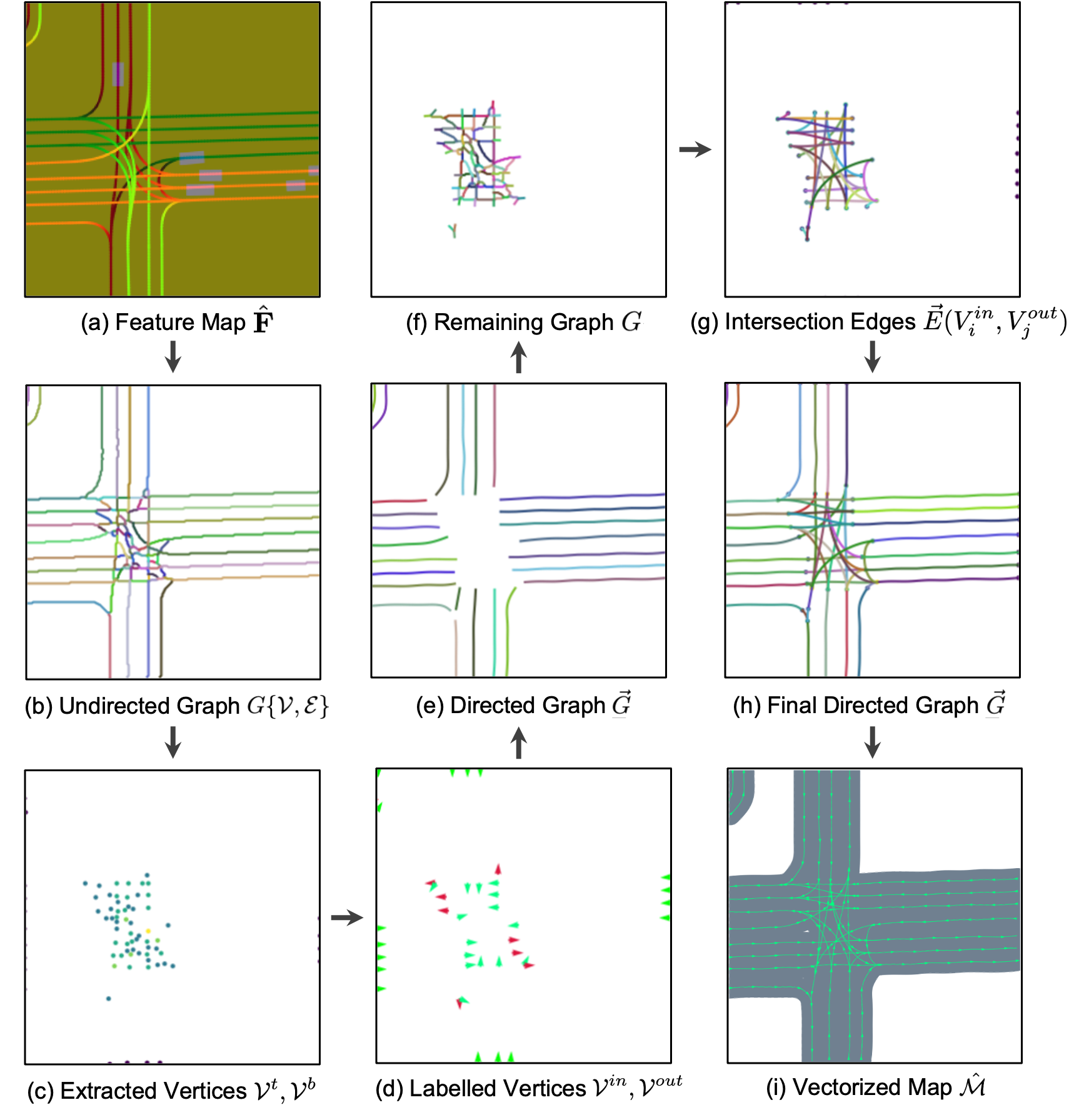}
    \caption{A visualization of the results obtained at each stage of the proposed vectorization method. The example feature map used represents a scene of 80$m\ \times$ 80$m$.}
    \label{figure-vectorization}
\end{figure}

To achieve this, we propose a graph-based algorithm that extracts lane geometries from the generated feature map and recovers their topological connections, as illustrated in Fig. \ref{figure-vectorization} and Algorithm \ref{algo-vector}. 
First, the original feature map is skeletonized into one-pixel thin features so that centerlines can be extracted from it as edges $\mathcal{E}$ based on their pixel values, and the locations where multiple centerlines intersect are detected as vertices $\mathcal{V}$ using criteria proposed in \cite{nodedetection1984}. 
Then, an undirected graph can be established as $G\{\mathcal{V}, \mathcal{E}\}$. 
Among all the vertices, we are particularly interested in the vertices with a degree of 1 (terminal vertices $\mathcal{V}^{t}$) and their corresponding neighbors (branching vertices $\mathcal{V}^{b}$). 
Next, the unit directional vectors $\Delta \Bar{p}$ at each vertex location can be recovered by inverting Eq. \ref{equation-dxdy}. 
All terminal vertices $\mathcal{V}^{t}$ and branching vertices $\mathcal{V}^{b}$ can be simply labeled into two types: entry or exit, based on their computed directions, as shown in green and red colors in Fig. \ref{figure-vectorization}(d). 
The assigned vertex type represents if an agent enters or exits an intersection along the driving direction. For each edge $E(V_{i}^{t}, V_{i}^{b})$ that connects a terminal vertex $V_{i}^{t}$ and its branching vertex $V_{i}^{b}$, a directed edge $\vec{E}$ can be extracted from graph $G$ into a new directed graph $\vec{G}$, following the driving direction. 
In the remaining of the undirected graph $G$, only intersection regions are present, including all branching vertices $\mathcal{V}^{b}$ with their labels, along with all the rest of the unclassified vertices. 
At this stage, for every pair of entry and exit vertices in the remaining graph $(V_{i}^{in}, V_{j}^{out})$, a b\'ezier curve $B_{i,j}$ is fitted to connect them if there exists a path $Path(V_{i}^{in}, V_{j}^{out})$ in the remaining of the undirected graph $G$. 
If the geometric of the fitted curve $B_{i,j}$ and the original path has a high Intersection over Union (IoU) score and the curvature of the fitted curve is within the permitted range, a new directed edge $\vec{E}(V_{i}^{in}, V_{j}^{out})$ is added to the directed graph $\vec{G}$. 
Finally, the extracted simple edges and the fitted edges at intersections are combined and converted to a set of polylines $\mathcal{P}$ to form the final vectorized map $\mathcal{\hat{M}}$. 

\subsection{Simulation}
The aim of the simulation stage is to generate a set of potential future trajectories $\hat{\Pi}_{i}$ for each generated agent $\hat{A}_{i}, i = \{1, \dots, N\}$ conditioned on the generated map context $\hat{\mathcal{M}}$ as well as the generated agents information $\hat{\mathcal{A}}$. A single future trajectory $\hat{\pi}_{i}$ belonging to a generated agent $\hat{A}_{i}$ can be represented as a sequence of states $\hat{s}_{i}^{t}$ at discrete time steps $t = \{1, \dots, \mathbf{T}\}$ within the time horizon $\mathbf{T}$. For each generated initial agent state $\hat{A}_{i}^{0}$, we wish to predict $K$ number of possible future candidate trajectories $\hat{\pi}_{i, k}, k = \{1, \dots, K\}$, as well as their probability $p_{i, k}$. With the simulation model $P_{sim}$, the simulation problem can be mathematically expressed as $\{ \hat{\pi}_{i, k}^{1:\mathbf{T}},p_{k} \}_{\substack{i\in\{ 1, \dots, N \}, k \in \{1, \dots, K\} }} = \mathcal{P}_{sim}\left( \hat{\mathcal{M}}, \{ \hat{A}_{1}^{0}, \dots, \hat{A}_{N}^{0} \} \right)$.
By observing the problem defined above, this conditional generation problem can be treated as a relaxed form of the trajectory prediction problem, which aims to predict the same set of future trajectories conditioned on agents' initial states with additional historical information over $\{-\mathbf{T}_{h}, \dots, 0\}$ time steps: $\mathcal{P}_{pred}\left( \hat{\mathcal{M}}, \{ \hat{A}_{1}^{-\mathbf{T}_{h}:0}, \dots, \hat{A}_{N}^{-\mathbf{T}_{h}:0} \} \right)$.
Thus, generic trajectory prediction methods can be employed in the simulation stage as the backbone. To demonstrate the usability of our generated data, we leverage the recent developments in the trajectory prediction field and repurposed an MTR\cite{mtr2023} model, which shows state-of-the-art multi-modal performance. The inputs to the model are configured to be the generated lane centerlines $\hat{\mathcal{M}} = \{ \hat{\mathcal{P}_{1}}, \dots, \hat{\mathcal{P}_{M}} \}$ and agents' initial states $\hat{\mathcal{A}} = \{ \hat{A}_{1}^{0}, \dots, \hat{A}_{N}^{0} \}$. From the same initial scene, the model's top $K$ marginal predictions $\{ \hat{\pi}_{i, k}^{1:\mathbf{T}},p_{k} \}$ are joined to form $K$ joint predictions $\{ \hat{\Pi}_{k} \mid k = {1, \dots, K} \}$, with each representing a distinct potential future scenario $\hat{D}_k$. 


\begin{figure*}[t]
    \centering
    \includegraphics[width=0.80\linewidth]{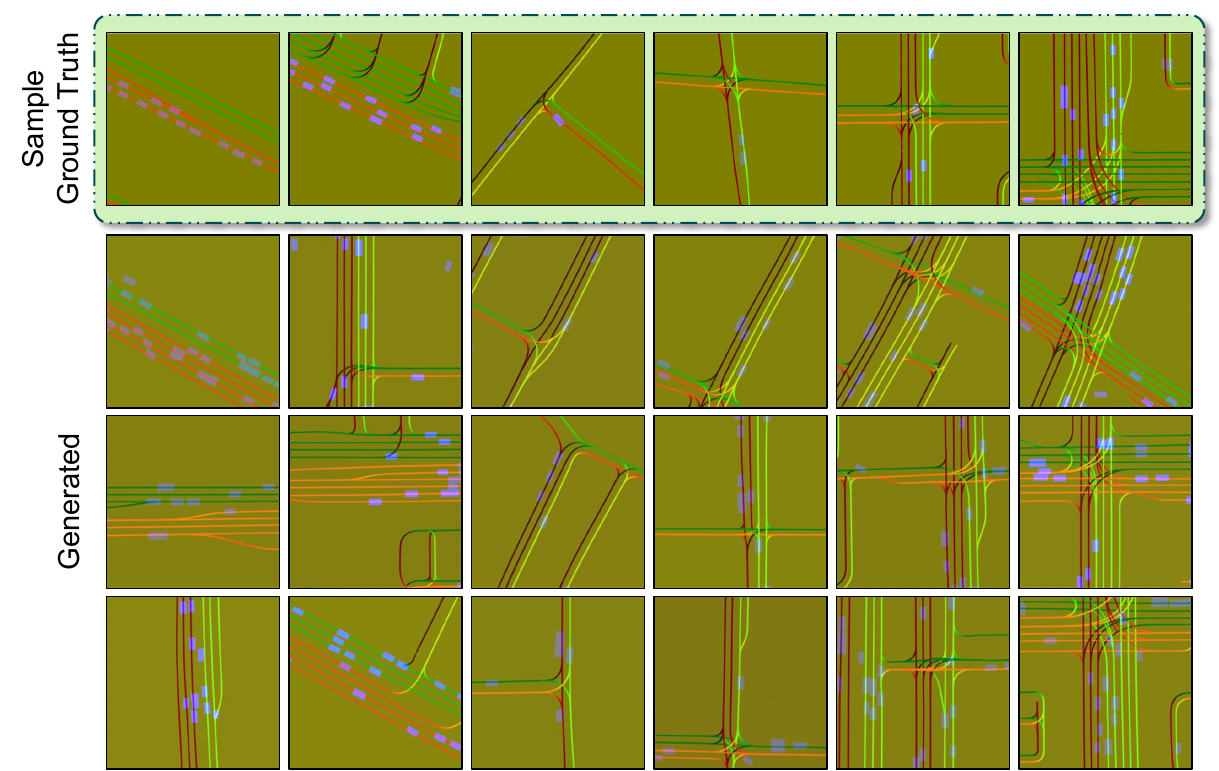}
    \caption{Qualitative evaluation of the raster initial scenes generated by \method. Compared to the ground truths, the generated samples show high fidelity in terms of road geometry, lane topology, and agents' initial state distribution. Meanwhile, the generated results demonstrate both inter-category and intra-category diversities.}
    \label{figure-diffusion-results}
\end{figure*}

\begin{table*}[ht]
\centering
\caption{Generation Fidelity \& Diversity}
\label{table-fid-div}
\begin{tabular}{ccccccccc}
    \hline
    \rowcolor[rgb]{0.863,0.863,0.863} {\cellcolor[rgb]{0.863,0.863,0.863}} & {\cellcolor[rgb]{0.863,0.863,0.863}} & \multicolumn{3}{c}{Channels} & {\cellcolor[rgb]{0.863,0.863,0.863}} & \multicolumn{3}{c}{Diversity} \\
    \rowcolor[rgb]{0.863,0.863,0.863} \multirow{-2}{*}{{\cellcolor[rgb]{0.863,0.863,0.863}}Task} & \multirow{-2}{*}{{\cellcolor[rgb]{0.863,0.863,0.863}}Size} & 1 & 2 & 3 & \multirow{-2}{*}{{\cellcolor[rgb]{0.863,0.863,0.863}}FID $\downarrow$} & Generated & Ground Truth & Discrepancy $\downarrow$ \\ 
    \hline
    \\[-1em]
    \multirow{2}{*}{Map-only} & \multirow{2}{*}{80$m$} & centerlines & centerlines & centerlines & 206.49 & 15.25 & 17.27 & 11.69$\%$ \\
     &  & {\cellcolor[rgb]{0.902,0.902,0.902}}dx & {\cellcolor[rgb]{0.902,0.902,0.902}}dy & {\cellcolor[rgb]{0.902,0.902,0.902}}gray & {\cellcolor[rgb]{0.902,0.902,0.902}}\textbf{35.75} & {\cellcolor[rgb]{0.902,0.902,0.902}}17.37 & {\cellcolor[rgb]{0.902,0.902,0.902}}17.76 & {\cellcolor[rgb]{0.902,0.902,0.902}}2.19$\%$ \\ 
    \hline
    \\[-1em]
    \multirow{4}{*}{Map + Agent} & \multirow{2}{*}{80$m$} & dx & dy & trajectories & 90.56 & 17.84 & 18.44 & 3.25$\%$ \\
     &  & {\cellcolor[rgb]{0.902,0.902,0.902}}dx & {\cellcolor[rgb]{0.902,0.902,0.902}}dy & {\cellcolor[rgb]{0.902,0.902,0.902}}initial states & {\cellcolor[rgb]{0.902,0.902,0.902}}46.81 & {\cellcolor[rgb]{0.902,0.902,0.902}}18.18 & {\cellcolor[rgb]{0.902,0.902,0.902}}18.41 & {\cellcolor[rgb]{0.902,0.902,0.902}}\textbf{1.25$\%$} \\
     & 120$m$ & dx & dy & initial states & 44.69 & 17.64 & 18.11 & 2.59$\%$ \\
     & 160$m$ & dx & dy & initial states & 46.40 & 17.63 & 17.91 & 1.56$\%$ \\
    \hline
\end{tabular}
\end{table*}

\section{EXPERIMENTS}
\label{section-experiments}

\subsection{Experiment Setup}
\label{section-experiments-setup}

We evaluate the performance of \method~on the Waymo Motion dataset \cite{Waymo2021}, which contains 70k driving scenarios of 20 seconds duration. Unless specified, all models are trained at a resolution of \(256\, \times 256\,\) on 4 NVIDIA V100 GPUs with a total batch size of 36 for 50 epochs. The training timesteps are set to 1000, and the AdamW optimizer is utilized with a learning rate of 1e-5. The number of model parameters for each is 56M. 

\subsection{Generation Fidelity \& Diversity} 

\subsubsection{Baselines}
To assess the variations in different generation methods, we explored two generation modes: one designed for scenarios featuring only static maps, and another for scenarios that combine both static maps and dynamic agents. 
For the map-only generation mode, two encoding strategies are explored: (1), which encodes centerlines with pixel values of $1.0$ (background as $0.0$) in all three feature map channels without the directional vector information, and (2), which encodes $dx$, $dy$ in the first two channels of the scenario with the third channel filled with $0.5$, following Eq. \ref{equation-dxdy}. 
For the map + agent generation mode, we experiment with two different encoding strategies: (3), which encodes the agents as a set of trajectories, and (4), which encodes the agents as a set of initial states. 

\subsubsection{Quantitative Evaluation}
Since the generated initial scenes are in a raster form, we employed the most commonly used metrics for raster data generation tasks to evaluate the quality of our generated results compared to the original dataset. 
To measure the \textbf{fidelity}, Fr\'echet Inception Distance (FID) \cite{FID2017} is used to compute the distance between the feature vectors of the 5$k$ generated initial scenes and the feature vectors of the 70$k$ initial scenes in the ground truth dataset, both extracted by the Inception V3 network. A lower FID score indicates a higher generation fidelity. To measure the \textbf{diversity}, the average feature distance \cite{DIV2019} among the feature vectors of the 5$k$ samples generated by each method is computed and compared with that of the 70$k$ ground truths in the Waymo dataset. The results are summarised in TABLE \ref{table-fid-div}.

Experimental results indicate that the method focusing only on the static map gives a higher FID score than generating initial scenes with both maps and agents. When compared with the method that only generates lane centerlines, our encoding strategy with directional vectors shows a significantly lower FID score, indicating an improvement in fidelity in the generated samples. It can also be concluded that, for agent encoding strategies, the one that encodes agents' initial states performs better than the one that encodes agent trajectories. It is observed that the ground truth data diversity across different representations differs slightly. Hence, the discrepancy in diversity between generated samples and the ground truth data is measured by percentage to provide a fair comparison. The results show that our encoding strategy with directional vectors improves the generated data diversity and is able to achieve a very similar level of diversity in data distribution to the ground truth. 

\subsubsection{Qualitative Evaluation}
To provide a qualitative comparison, several typical categories of driving scenarios are selected from both the ground truth dataset and the samples generated by our method, as shown in Fig. \ref{figure-diffusion-results}. It can be observed that samples generated by \method~are realistic enough when compared to the ground truths in terms of road geometry, connectivity logic, and agents' initial state distribution. Additionally, our method is able to learn a wide range of scenario styles and generate results with sufficient diversity both within each category and across multiple categories. 

\begin{figure*}[ht]
    \centering
    \includegraphics[width=0.80\linewidth]{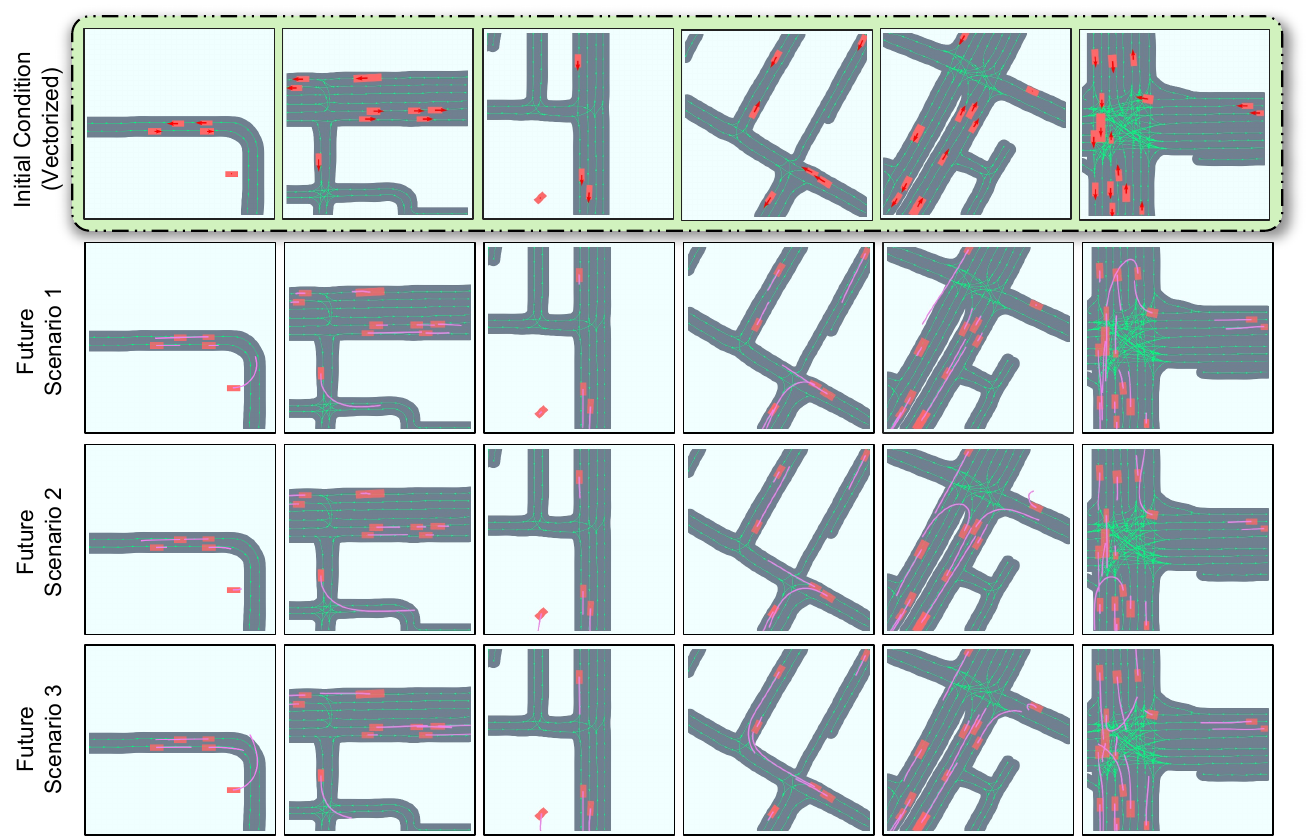}
    \caption{Qualitative evaluation of the different possible future scenarios generated from the same initial scene. Comparing the scenarios in each column, each future scenario presents distinct agent behaviors starting from the same initial scene.}
    \label{figure-simulation-results}
\end{figure*}

\subsubsection{Ablation Study} 
The fidelity and diversity scores of various ablated versions of our method are presented in TABLE \ref{table-fid-div}. While maintaining a constant input resolution, rectangular areas of different sizes are extracted from ground truth scenarios to train and evaluate each ablated version. The fidelity and diversity results indicate that \method~is not sensitive to the changes in the range of the scenario. 

\subsection{Vectorization Fidelity} 

\subsubsection{Baselines}
Since the generated initial scenes are in customized representations, two other baseline methods are proposed and compared to evaluate the performance of our vectorization pipeline: (1) a Transformer-based vectorization model and (2) a graph-based algorithm without curve fitting. 

In our Transformer-based vectorization model, a variant of the DETR\cite{carion2020end} detector is designed for extracting lane vectors from the BEV feature maps. In line with the method described in LaneGAP\cite{lanegap}, the initial raw ground truth polylines are assembled as a path that traverses the map from one edge to another. A hierarchical query is also implemented for the transformer decoder, explicitly encoding centerline elements for enhanced accuracy\cite{MapTR}.

\subsubsection{Quantitative Evaluation} Two graph-based metrics are selected to assess the fidelity of vectorization, namely GEO and TOPO metric \cite{he2022lane}. The GEO-metric compares the geometric accuracy and recall between the vectorized results and the ground truth, while the TOPO-metric further considers the topological connectivity as well. The vectorization process is evaluated on the Waymo dataset since the generated initial scenes do not have their corresponding ground truth data. For each ground truth feature map $\mathbf{F}$, we predict the vectorized data $\mathcal{M}_{pred}$ and compare it with the ground truth data $\mathcal{M}$. 

\textbf{GEO-metric}: For the ground truth graph $G$, and extraction result $\hat{G}$, two vertices are paired if their distance is smaller than a given threshold, and paired extraction results are noted as $\hat{V}_{match} \in \mathcal{\hat{V}}_{match}$. Then, GEO precision and recall are computed as follows:
\begin{equation}
\label{equation-geo-metric}
    \begin{aligned}
        \text{Pre}_{\text{GEO}}=\frac{\left|\mathcal{\hat{V}}_{\text {match }}\right|}{|\hat{\mathcal{V}}|}, \ 
        \text{Rec}_{\text{GEO}}=\frac{\left|\mathcal{\hat{V}}_{\text {match }}\right|}{|\mathcal{V}|} 
    \end{aligned}
\end{equation}

\textbf{TOPO-metric}: Different from the GEO-metric, which only considers geometric differences without considering the topological connectivity, a second metric called TOPO metric is employed. After the GEO pairing, each vertex pair builds a subgraph $(S_V,\hat{S}_{\hat{V}})$ based on the vertex connectivity. Each subgraph will include neighbor vertices on the path by distance $r$ on each side. Then, the GEO score between $(S_V, \hat{S}_{\hat{V}})$ is computed. Finally, the TOPO score is calculated among all subgraphs and could be considered as GEO metrics weighted by subgraphs, as defined in Eq. \ref{equation-topo-metric}.

During the evaluation, each path is interpolated to intervals of 0.5 meters. For the ground truth graph $G$ and the extraction result $\hat{G}$, two vertices are paired if their distance is less than 1.5 meters, and the subgraph is established within a distance of $r = 50$ meters. Experiments are conducted on 1$k$ ground truth examples, and results are listed in TABLE \ref{table-vectorization}. Results indicate that the Graph-fitting method has the highest vectorization fidelity. The graph-based approach outperforms the DETR-path method as it directly extracts polylines from the pixels. Furthermore, the graph-based method exhibits less degradation in performance as the scenario size increases. The Graph-fitting method excels in accurately fitting lane shapes, resulting in a significant improvement in the Precision scores. By comparing the results of different map ranges, a recommended range of 80 meters is concluded to strike a balance between scenario size and vectorization accuracy.

\begin{equation}
\label{equation-topo-metric}
    \begin{aligned} 
    \text{Pre}_{\mathrm{TOPO}} & =\frac{\sum_{\text{matched }(v, \hat{v})} \operatorname{Prec}_{\mathrm{GEO}}\left(S_{v}, \hat{S}_{\hat{v}}\right)}{|\hat{V}|} \\
    \text{Rec}_{\mathrm{TOPO}} & =\frac{\sum_{\text{matched }(v, \hat{v})} \operatorname{Rec}_{\mathrm{GEO}}\left(S_{v}, \hat{S}_{\hat{v}}\right)}{|V|}
    \end{aligned}
\end{equation}

\begin{table}[htbp]
\centering
\caption{Fidelity of Vectorized Samples}
\label{table-vectorization}
\begin{tabular}{cccccccc} 
    \hline
    \rowcolor[rgb]{0.863,0.863,0.863} {\cellcolor[rgb]{0.863,0.863,0.863}} & {\cellcolor[rgb]{0.863,0.863,0.863}} & \multicolumn{3}{c}{GEO Metric~$\uparrow$} & \multicolumn{3}{c}{TOPO Metric~$\uparrow$} \\
    \rowcolor[rgb]{0.863,0.863,0.863} \multirow{-2}{*}{{\cellcolor[rgb]{0.863,0.863,0.863}}Method} & \multirow{-2}{*}{{\cellcolor[rgb]{0.863,0.863,0.863}}Size} & Pre & Rec & F1 & Pre & Rec & F1 \\ 
    \hline
    \multirow{3}{*}{DETR-path~} & 40$m$ & 0.64 & 0.81 & 0.72 & 0.56 & 0.57 & 0.56 \\
     & 80$m$ & 0.52 & 0.67 & 0.59 & 0.37 & 0.33 & 0.35 \\
     & 120$m$ & 0.41 & 0.53 & 0.46 & 0.22 & 0.18 & 0.2 \\ 
    \hline
    \multirow{3}{*}{Graph} & 40$m$ & 0.7 & 0.86 & 0.77 & 0.59 & \textbf{0.75} & 0.66 \\
     & 80$m$ & 0.65 & 0.78 & 0.71 & 0.5 & 0.63 & 0.56 \\
     & 120$m$ & 0.64 & 0.56 & 0.6 & 0.47 & 0.37 & 0.42 \\ 
    \hline
    \rowcolor[rgb]{0.902,0.902,0.902} {\cellcolor[rgb]{0.902,0.902,0.902}} & 40$m$ & \textbf{0.95} & \textbf{0.86} & \textbf{0.91} & \textbf{0.92} & 0.67 & \textbf{0.77} \\
    \rowcolor[rgb]{0.902,0.902,0.902} {\cellcolor[rgb]{0.902,0.902,0.902}} & 80$m$ & 0.92 & 0.85 & 0.88 & 0.88 & 0.56 & 0.68 \\
    \rowcolor[rgb]{0.902,0.902,0.902} \multirow{-3}{*}{{\cellcolor[rgb]{0.902,0.902,0.902}}\begin{tabular}[c]{@{}>{\cellcolor[rgb]{0.902,0.902,0.902}}c@{}}Graph + \\Curve Fitting\end{tabular}} & 120$m$ & 0.9 & 0.82 & 0.86 & 0.86 & 0.47 & 0.6 \\
    \hline
\end{tabular}
\end{table}%

\subsection{Simulation Results}


\subsubsection{Qualitative Evaluation}
For this experiment, the simulation model is trained on the same Waymo Motion dataset. Based on the experiment results in the previous section, the scenario size is set as 80 meters. For each generated and vectorized initial scene $\{ \mathcal{\hat{M}}, \mathcal{\hat{A}} \}$, the top 3 future trajectories are generated by the simulation model and visualized in Fig. \ref{figure-simulation-results}. By comparing the scenarios in each column, it can be observed that each future scenario presents distinct agent behaviors starting from the same initial scene.


\section{CONCLUSIONS \& DISCUSSION} 
In this paper, we present the first method that learns to generate driving scenarios from the collected real-world driving dataset from scratch. Our method demonstrates that the task of generating novel driving scenarios that comply with real-world distributions is possible. Experimental results highlight the capability of \method~in generating quality, diversity, and scalability on real-world datasets. 
We believe that such capability can potentially benefit the development of data-driven methods in autonomous driving and have unlimited prospective applications outside robotics. 
Since the generation of an entire dynamic scenario is a new task, there are currently no quantitative metrics to directly compare the distribution of the generated dynamic scenarios consisting of both maps and agent behaviors to the ground truth distribution. We wish to investigate potential metrics as part of our future work.


\addtolength{\textheight}{-3cm}   






\bibliographystyle{IEEEtran}
\bibliography{root}

\end{document}